\documentclass{article}
\pdfoutput=1
\usepackage{fullpage}
\usepackage[utf8]{inputenc} 
\usepackage[T1]{fontenc}    
\usepackage{hyperref}       
\usepackage{url}            
\usepackage{booktabs}       
\usepackage{amsfonts}       
\usepackage{nicefrac}       
\usepackage{microtype}      
\usepackage{graphicx}
\usepackage{amsmath}
\usepackage{bm}
\usepackage{amsthm}
\usepackage{mathtools}
\usepackage[mathscr]{euscript}
\usepackage{bbm}
\usepackage{float}
\usepackage{subfig}
\usepackage[ruled,vlined]{algorithm2e}

\newcommand{\set}[2][]{#1 \{ #2 #1 \} }
\newcommand{\ignore}[1]{}

\makeatletter
\newtheorem*{rep@theorem}{\rep@title}
\newcommand{\newreptheorem}[2]{%
\newenvironment{rep#1}[1]{%
 \def\rep@title{#2 \ref{##1}}%
 \begin{rep@theorem}}%
 {\end{rep@theorem}}}
\makeatother

\newtheorem{theorem}{Theorem}
\newreptheorem{theorem}{Theorem}
\newtheorem{lemma}{Lemma}
\newreptheorem{lemma}{Lemma}

\hypersetup{
  colorlinks   = true,
  urlcolor     = blue,
  linkcolor    = blue,
  citecolor   = blue
}

\title{A Weighted K-Center Algorithm for Data Subset Selection
}
\author{
Srikumar Ramalingam \qquad Pranjal Awasthi \qquad Sanjiv Kumar \vspace{0.1cm}
\\
Google Research, New York\\
{\tt\small \{rsrikumar, pranjalawasthi, sanjivk\}@google.com}
}

\date{}

\begin{document}

\maketitle

\begin{abstract}
The success of deep learning hinges on enormous data and large models, which require labor-intensive annotations and heavy computation costs. Subset selection is a fundamental problem that can play a key role in identifying smaller portions of the training data, which can then be used to produce similar models as the ones trained with full data. Two prior methods are shown to achieve impressive results: (1) margin sampling that focuses on selecting points with high uncertainty, and (2) core-sets or clustering methods such as $k$-center for informative and diverse subsets. We are not aware of any work that combines these methods in a principled manner. To this end, we develop a novel and efficient factor 3-approximation algorithm to compute subsets based on the weighted sum of both $k$-center and uncertainty sampling objective functions. To handle large datasets, we show a parallel algorithm to run on multiple machines with approximation guarantees. The proposed algorithm achieves similar or better performance compared to other strong baselines on vision datasets such as CIFAR-10, CIFAR-100, and ImageNet.
\end{abstract}

\section{Introduction}
We are witnessing ground-breaking results of deep learning in various domains such as computer vision~\cite{Krizhevsky2012,Szegedy2015,He2016DeepRL}, speech~\cite{Hinton2012}, and natural language processing~\cite{Devlin2018}. This makes us ponder the key factors behind this revolution: is it the availability of the large datasets, the actual learning algorithms, or both? ML models rely on very deep networks and enormous labeled datasets, requiring exorbitant computational and human labeling efforts. For example, the market for data annotation costs have crossed one billion US dollars in 2020, and it is estimated to hit seven billion in 2027. Human annotation of semantic segmentation labels takes about 45-60 minutes~\cite{badrinarayanan2010label} for a single image. 

In most vision and NLP applications, unlabeled data is unlimited and is usually available at no cost. To directly reduce the human annotation costs, this paper focuses on identifying smaller subsets of training data that can lead to accurate models with marginal or no loss in performance compared to the ones trained on the full dataset. 
Among the several approaches for subset selection, two 
are shown to achieve impressive results: (1) the classical margin sampling algorithm that selects points based on the uncertainty in the class prediction scores~\cite{Roth2006MarginBasedAL}, and (2) the $k$-center clustering algorithm~\cite{sener2017active} based on core sets. One may wonder about the natural extension of these two powerful algorithms: is there a principled method that jointly uses both these measures for computing more informative subsets? In fact, a recent work on $k$-center~\cite{sener2017active} states this as an open problem with potential benefits. 

The real challenge in combining the objective functions arises from the $k$-center method. The margin sampling is an exact algorithm since the subset selection is essentially about picking the top $k$ points based on the uncertainty values. However, $k$-center clustering is NP-hard, and comes with a greedy algorithm with a factor 2-approximation guarantee~\cite{gonzalez1985clustering}. Any algorithm that optimizes the weighted sum of these two objective functions will also be an approximate one, and the interesting research lies in finding a method with constant approximation guarantees. 
To this end, we propose a novel and principled approach to combine these methods, and prove that our algorithm comes with factor 3-approximation guarantee, which is not far from the 2-approximation guarantee for the standard $k$-center objective.  

Our work has similarities to and certain crucial differences from the classical setting of active learning (See~\cite{settles2009active} for a detailed survey). 
In active learning one 
uses an iterative approach where a single point is labeled and used to update the model in each iteration. In general, the use of a single sample at a time is very ineffective both in cost-effectiveness of employing the human annotator for just labeling one sample at a time, and the additional overhead in making a negligible improvement to the model. Instead, we focus on choosing a single large subset of the given training data. This is reminiscent of batch active learning~\cite{Amin2020UnderstandingTE,Shui2020DeepAL,Kim2021TaskAwareVA,ghorbani2021data} except that we do not perform multiple iterations of the selection procedure. In particular, we study single-shot setting where we consider one budget or batch and the performance is studied for different subset selection algorithms.

We use an initial model by training with a small random subset of 10\% of the annotated data. We will use this initial model to generate features (embeddings) and class probability scores that will be used for computing the subsets. We will evaluate the quality of these subsets by training deep networks using these subsets, and evaluating the models on test sets. 

We summarize the contributions of this paper below: 
\begin{itemize}
\item We design an efficient and novel algorithm to minimize the weighted sum of $k$-center and margin sampling objective functions for subset selection. 
\item We prove that the proposed weighted $k$-center algorithm achieves a constant 3-factor approximation guarantee.
\item We propose an alternate parallel algorithm that can utilize multiple machines, and prove that the algorithm comes with constant 14-factor approximation guarantee.
\item We outperform other subset selection algorithms on standard vision datasets such as CIFAR-10, CIFAR-100, and ImageNet~\cite{krizhevsky2009learning, russakovsky2014imagenet}. 
\end{itemize}

\section{Previous Work}
{\bf Weighted $k$-Center.}
The $k$-center clustering problem has a long and rich history with applications in multiple domains. From a theoretical perspective the problem admits a greedy $2$-approximation \cite{gonzalez1985clustering, hochbaum1985best}. Parallel algorithms in the map reduce framework have also been studied recently for the $k$-center objective~\cite{ene2011fast, im2015fast, malkomes2015fast, mcclintock2016efficient}. Different notions of weighted $k$-center have also been studied in the from an approximation algorithms perspective. However these differ from the formulation that we consider. The work of~\cite{Kumar2016CapacitatedKP} studies the version where each vertex has a weight, and the cost of assigning a point to a center is the distance times the weight of the center. Here the goal is to find a set of $k$ centers that minimizes the maximum weighted distance. Following a long line of work~\cite{kariv1979algorithmic, megiddo1983new, jeger1985algorithms}, the work of Chen and Wang~\cite{wang1990heuristic} provided a $2$-approximation algorithm for this formulation. This is the best possible since the unweighted $k$-center objective itself is NP-hard to approximate to a $2-\epsilon$ factor for any constant $\epsilon > 0$~\cite{hsu1979easy}. Extensions of the above formulation where a vertex can only serve as a center for a limited number of points have also been studied~\cite{kumar2016capacitated}. However, the approximation guarantees in such settings are much worse and depend on $n$, the number of data points. Another formulation of weighted $k$-center concerns the setting where each vertex has a weight and there is an upper bound on the weight of the set of centers that can be chosen. Constant factor approximation algorithms are known in this case~\cite{wirth2005approximation}.

{\bf Active learning.}
Uncertainty sampling is one of the simplest and most effective methods, where the strategy is to select most challenging or uncertain points first, and thereby avoid the need to label the easier ones later. In the classification setting, class probability scores from the model can be used to identify uncertain samples. Popular choices include the use of top class probability, difference between the best and second best class probabilities~\cite{Lewis1994,Scheffer2001}, entropy measure~\cite{Holub2008,Joshi2009}, and the geometric distance to the decision boundary~\cite{Tong2002,Brinker2003}. Other active learning methods include disagreement based approaches where we train several models and look for disagreement on the unlabeled data~\cite{Gilad-Bachrach2005,Seung1992}. Such uncertainty-based methods for subset selection are shown to suffer from sampling bias~\cite{Schutze2006}, and a hierarchical clustering of the dataset can be used to guide the sampling. In many problems, a single active learning strategy may not work well, and different strategies can be combined by formulating this as a multi-arm bandit problem~\cite{Hsu2015}. Theoretical analysis in active learning is focused on obtaining label complexity bounds~\cite{Cohn1994} by analyzing the shrinkage of version space~\cite{Dasgupta2008} with each label query. There has also been extensive work on active learning in non-parametric settings where certain low-noise assumptions are required to avoid the curse of dimensionality~\cite{castro2008minimax, hanneke2011rates, locatelli2017adaptivity}.


{\bf Submodular subset Selection.} Several utility functions involving diversity and representativeness for subset selection are shown to be submodular, which is a discrete analogue of convex functions~\cite{Nemhauser_MP1978,Kolmogorov2004, ramalingam2017dam}. The task of subset selection is usually formulated as maximization of submodular functions. While the maximization is NP-hard, efficient greedy algorithms with constant factor approximation guarantees have been used in many applications beyond data selection~\cite{Carbonell98,Simon2007,krause2014submodular,golovin2010adaptive,Wei2015,lin-bilmes-2011-class,kaushal2019learning,Prasad2014,Kim2016,Prasad14, golovin2010adaptive,RamalingamRN17, Golovin2011,Kim2016,ramalingam2021balancing,Kothawade2021,Kaushal2021}. 

{\bf Clustering.}
$K$-medoids is a clustering algorithm that selects $k$ centers by minimizing the sum of distances from each of the $k$ cluster centers to the points belonging to the cluster~\cite{kaufmanl1987clustering,Park2009}. $K$-medoids has been used for prototype learning, where the goal is to identify subset of training points that best represent the entire dataset. Subset selection using clustering has been used before to identify redundant data points in training data, and in particular, 10\% of the images in standard vision datasets such as CIFAR-10 and ImageNet are shown to be redundant~\cite{birodkar2019semantic}. There is also a large body of work that focuses on designing polynomial time approximation algorithms for various clustering objectives such as $k$-means and $k$-median~(see~\cite{awasthi2014center} for a survey.)

\section{Preliminaries and notations}
We consider a classification setting with $L$ labels and unlabeled training data given by $\set{U} = \left\{ x_1, \dots, x_n \right\}$. We will use $G=\{1,\dots,n\}$ to denote the indices in the dataset and let $y_i \in \set{Y}$ denote the ground truth label corresponding to $x_i$. We will denote the trained network by the function $\psi: \set{X} \rightarrow \mathbb{R}^L $ that takes an input data point and outputs a probability vector over $L$ classes. The network $\psi: \set{X} \rightarrow \mathbb{R}^L $ can usually be  treated as a composition of two functions: an {\it embedding function} $\phi: \set{X} \rightarrow \mathbb{R}^E$ and a {\it discriminator function} $h: \mathbb{R}^E \rightarrow \mathbb{R}^L$. The former maps the input to and $E$-dimensional feature, and the latter maps the embedding to output predictions. 

\noindent
{\bf Seed model:}
We are given a few data points with labels to build our initial model. We refer to this data as seed $(x_i,y_i); i \in \set{I}$, and we denote the initial model as $\psi^{\set{I}}$. This model will be used to compute features and class prediction scores for all the unlabelled data. 

\noindent
{\bf Nearest neighbor graph.}
Both the standard $k$-center and our weighted $k$-center algorithms depend on distances between data points. The distances are computed based on the embeddings or the features of the data points. In particular, we build a nearest neighbor graph $(G,E)$, where $E$ denotes the set of edges associated with nearby data points. 

\noindent
{\bf Uncertainty sampling.} Each data point is associated with a weight based on its uncertainty. Specifically, the data point has a higher weight if it is least uncertain, and vice versa. There are three popular choices for uncertainty: least confident, margin~\cite{Scheffer2001}, and entropy. In this paper, we choose the margin score, which is applicable in a multi-class setting, and gives preference to ``hard to classify" or ``low margin" examples (see~\cite{Scheffer2001} for more details). More specifically, the weight of a point $i$ is given by:
\begin{equation}
w(i) = P(Y = bb | x_i) - P(Y = sb | x_i),
\label{eq:margin_sampling}
\end{equation}
where $bb$ and $sb$ denote the best and the second best predicted labels for $x_i$ according to our initial model $\psi^{\operatorname{seed}}$, and $P(Y = \cdot | x_i)$ denotes the class probabilities predicted by the same model for $x_i$. 

\noindent
{\bf The $k$-center problem.} Here we compute a subset $S$ of $k$ centers from a large set of points $G$, such that the maximum distance of any point in $G$ to its closest center in $S$ is minimized~\cite{Hochbaum1986}. Specifically, we minimize the following objective function to compute $S$:
\begin{align}
    \label{eq:k-center-obj}
    \max_{i \in G} \min_{j \in S} d(i,j)
\end{align}

\section{Weighted $k$-Center}
We next propose the weighted $k$-center objective that will be used to combine the clustering based active learning with margin based approaches. Let $G$ be a set of data points, and for $i,j \in G$, define $d(i,j)$ to be the distance between the two points. Furthermore, denote by $w(i)$ the weight of data point $i$. Given a parameter $\lambda$, the weighted $k$-center problem asks for a set $S \subseteq G$ that minimizes:
\begin{align}
    \label{eq:weighted-k-center-obj}
    \max_{i \in G} \min_{j \in S} d(i,j) + \lambda w(S),
\end{align}
where
\begin{align*}
    w(S) = \sum_{i \in S} w(i).
\end{align*}\label{sec:weighted-k-center}

\begin{figure*}[!t]
    \centering
        \includegraphics[width=0.80\textwidth]{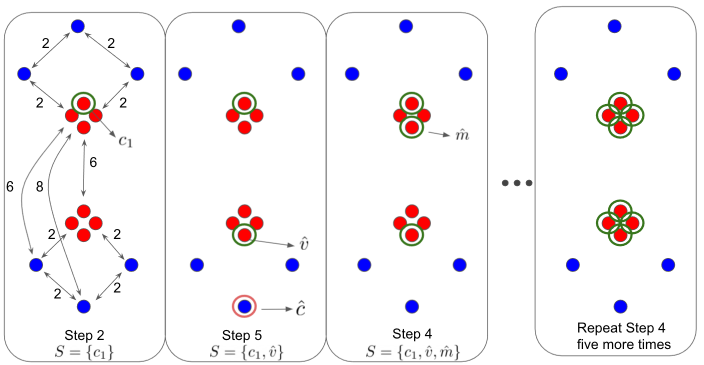}
    \caption{\small \it In this example, we have a total of 14 points and we select $k=8$ points. The red points have 0.5 weights and the blue ones have 1. The optimum $k$-center algorithm will select all the blue points and one red point each from the two red clusters yielding a total cost of 1. While this solution enforces diversity, it completely ignores the weights and obtains a solution with a total weight of 7. Let us consider the weighted k-center case where $\lambda=1$. The optimum solution for the weighted k-center is a subset with 8 points from both the red clusters yielding a total cost of 6 (4 from weight and 2 from distance). Hence the optimum value of gamma=2. Consider the invocation of Algorithm 1 with this value of gamma. In step 2, we select a sample with the lowest weight as $c_1$ and without loss of generality we choose a point from one of the red clusters. Next, we use step 5 to choose $\hat{c}$ that is at least $3\gamma = 6$ from $|S|$, and consequently a lowest weight sample $\hat{v}$ in the $\gamma=2$ neighborhood of $\hat{c}$. Next, we find 6 points based on the weights using step 4. The selected subset is shown with green circles, and we manage to find the optimum. }
    \label{fig:schematic}
\end{figure*}

Notice that when $\lambda = 0$, the above objective reduces to the standard $k$-center objective. We next discuss come challenges in designing an efficient approximation algorithm for the weighted objective. The greedy algorithm for standard $k$-center~\cite{gonzalez1985clustering} works by iteratively building the set $S$ by greedily picking the farthest point from $S$ and adding it as the next choice of the center. However, in the presence of weighted points there is no guarantee that the above procedure will return a solution of small weight. In particular, the set returned by the  greedy algorithm can be arbitrarily bad in the worst case. In order to overcome this we propose an alternate approach in Algorithm~\ref{alg:weighted-k-center-improved}. Our proposed algorithm also builds the set $S$ of centers iteratively. However, the key difference is that at each time step, we do not simply pick a point that is the farthest from $S$, instead pick a point of minimum weight among the points that are suitably far from $S$. However, this b itself is not sufficient to guarantee a good approximation, as there may be case where only a few select points in the dataset have small weights, and they in turn also minimize the $k$-center  part of the objective in Eq. \eqref{eq:weighted-k-center-obj}. To deal with such cases, we again pick a ball of a suitable radius around the current guess $c$, and pick the point $\hat{c}$ of minimum weight as our choice for the next center to be added to $S$. We next formally present our method in Algorithm~\ref{alg:weighted-k-center-improved} and the associated guarantee in Theorem~\ref{thm:weighted-k-center-improved}. We also show a simple example in Fig.~\ref{fig:schematic} to illustrate the different steps in Algorithm~\ref{alg:weighted-k-center-improved}.


\begin{algorithm}
\SetAlgoLined
 1. \textbf{Input:} Graph $G = (V, V \times V)$, hyper-parameter $\gamma (\lambda)$.\\
 2. Initialize $S = \{c_1\}$ where $c_1$ is the data point with minimum weight.\\
 3. While $|S| < k$ do:
 \begin{itemize}
     \item 4. If all points are within $3\gamma (\lambda)$ from the current $S$ then append to $S$ the data point of minimum weight $\hat{m}$ among the remaining points.
     \item 5. Otherwise, among the points that are more than $3\gamma (\lambda)$ away from $S$, let $\hat{c}$ be the one with the minimum weight. Add to $S$ the vertex $\hat{v}$ of minimum weight in the $\gamma (\lambda)$-neighborhood of $\hat{c}$.
 \end{itemize}
 6. Output $S$.
 \caption{An improved approximation algorithm for the weighted $k$-center objective.}\label{alg:weighted-k-center-improved}
\end{algorithm}

\begin{theorem}
\label{thm:weighted-k-center-improved}
For a graph $G = (V, V \times V)$, let $S^*$ be the subset that optimizes the $k$-center objective in \eqref{eq:weighted-k-center-obj}. Then there exists a threshold $\gamma (\lambda)$ such that Algorithm~\ref{alg:weighted-k-center-improved} when run with $\gamma (\lambda)$ outputs a set $S$ of size $k$ such that 
\begin{align*}
    &\max_{i \in G} \min_{j \in S} d(i,j) +  \lambda \sum_{\hat{c}_j \in S} w(\hat{c}_j)\\ &\leq  
    3 \Big(\max_{i \in G} \min_{j \in S^*} d(i,j) + \lambda \sum_{c^*_i \in S^*} w(c^*_i) \Big).
\end{align*}
\end{theorem}
\begin{proof}
Let $C^*_1, C^*_2, \dots, C^*_k$ be the optimal clustering according to the objective in Eq. \eqref{eq:weighted-k-center-obj}. Furthermore, let $c^*_1, c^*_2, \dots, c^*_k$ be the corresponding cluster centers and therefore the optimal set $S^*$ according to \eqref{eq:weighted-k-center-obj} corresponds to $S^* = \{c^*_1, c^*_2, \dots, c^*_k\}$. Define $\gamma (\lambda)$ to be the clustering cost or the k-center objective of using the set $S^*$, i.e., 
$$
\gamma (\lambda) \coloneqq \max_{i \in G} \min_{c_j \in S^*} d(i, c_j).
$$
We will consider the invocation of Algorithm~\ref{alg:weighted-k-center-improved} with this $\gamma (\lambda)$. We will show that every center $\hat{c}_i$ picked in step 5 can be matched to a unique optimal center $c^*_i$ such that $w(\hat{c}_i) \leq w(c^*_i)$ and using $\hat{c}_i$ as center for the points in $C^*_i$ has a cost of at most $3\gamma (\lambda)$. We will call an optimal cluster as {\em covered} if one of the picked centers in step 5 has been matched to the center of the given cluster. We first consider step 2 where the center $c_1$ with the minimum weight is added. Without loss of generality assume that $c_1$ belongs to the optimal cluster $C^*_1$. Then we can match $c_1$ to $c^*_1$ and using $c_1$ as center to cluster points in $C^*_1$ has a cost of at most $2\gamma (\lambda)$. Furthermore, notice that $w(c_1) \leq w(c^*_1)$.

Next consider a particular invocation of step 5 of the algorithm. Since all points in covered clusters are within $3\gamma (\lambda)$ of a center in $S$, the new point $c_i$ must belong to an uncovered optimal cluster, say $C^*_i$. Notice that any point in $C^*_i$ is at most $2\gamma (\lambda)$ from $c_i$ and as a result is at most $3\gamma (\lambda)$ away from $\hat{c}_i$. Hence we can use $\hat{c}_i$ to cover the cluster $C^*_i$. Furthermore, $w(\hat{c}_i) \leq w(c^*_i)$ holds true since $c^*_i$ is within $\gamma (\lambda)$ distance of $c_i$ and hence would have been picked instead of $\hat{c}_i$ if it had lower weight. Finally, notice that if step 4 is ever executed then all the remaining points are already within a distance of $3\gamma (\lambda)$ from an existing center, and adding points of minimum weight among the remaining points will have a lower total weight than adding the optimal centers of the remaining uncovered clusters. As a result we obtain a set $S$ of size $k$ such that
\begin{align*}
&\max_{i \in G} \min_{\hat{c}_j \in S} d(i, \hat{c}_j) \leq 3\gamma (\lambda)\\
&\sum_{\hat{c}_j \in S} w(\hat{c}_j) \leq \sum_{c^*_i \in S^*} w(c^*_i).  
\end{align*}
Combining the above we get that the set $S$ achieves a $3$-approximation to the objective in Eq. \eqref{eq:weighted-k-center-obj} as shown below:
\begin{align*}
    &\max_{i \in G} \min_{j \in S} d(i,j) +  \lambda \sum_{\hat{c}_j \in S} w(\hat{c}_j) \\
    &\leq 3 \Big(\max_{i \in G} \min_{j \in S^*} d(i,j) + \lambda \sum_{c^*_i \in S^*} w(c^*_i) \Big).
\end{align*}

\end{proof}
Algorithm~\ref{alg:weighted-k-center-improved} requires knowledge of the parameter $\gamma (\lambda)$ that corresponds to the $k$-center cost incurred by the optimal solution to the objective in Eq. \eqref{eq:weighted-k-center-obj}. We next present a simple lemma stating that the true value of $\gamma (\lambda)$ always lies between the optimal $k$-center objective value~(i.e., $\lambda=0$), and the $k$-center objective value incurred by the solution that simply picks the $k$ centers of minimum weight~(i.e., $\lambda = \infty$). The Lemma below provides a useful range of $\gamma (\lambda)$ values to search from.

\begin{lemma}
\label{lem:range-of-gamma}
Let $\gamma (\lambda)$ be the $k$-center objective value incurred by the optimal solution for Eq. \eqref{eq:weighted-k-center-obj}. Furthermore, let $\gamma_1$ be the $k$-center objective value obtained by setting $\lambda=0$ in Eq. \eqref{eq:weighted-k-center-obj}, and let $\gamma_2$ be the $k$-center objective value attained by the clustering obtained by setting $\lambda=\infty$ in Eq. \eqref{eq:weighted-k-center-obj}. Then we have that $\gamma_1 \leq \gamma (\lambda) \leq \gamma_2$. 
\end{lemma}
\begin{proof}
It is easy to see that $\gamma_1 \leq \gamma (\lambda)$, since $\gamma_1$ is the minimum $k$-center objective value achievable by any $k$-clustering. To see the other direction let $S_1 = \{c^*_1, \dots, c^*_k\}$ be the centers picked by the optimal solution to Eq. \eqref{eq:weighted-k-center-obj}, and let $S_2 = \{c_1, \dots, c_k\}$ be the centers picked when $\lambda = \infty$. Then we have that $w(S_2) \leq w(S_1)$. Hence, if $\gamma (\lambda) > \gamma_2$, then $\{c^*_1, \dots, c^*_k\}$ cannot be the optimal solution since we will have that $\gamma_2 + \lambda w(S_2) < \gamma (\lambda) + \lambda w(S_1)$, thereby leading to a contradiction.
\end{proof}

Notice that our algorithm relies on a hyperparameter $\gamma(\lambda)$. It is important to consider cases when we do not have an accurate estimate of this quantity. Our approximation guarantees presented in Theorem~\ref{thm:weighted-k-center-improved} scale gracefully on the amount of over-estimation factor. In particular of Algorithm~\ref{alg:weighted-k-center-improved} is run with $\gamma' = \alpha \gamma(\lambda)$ for $\alpha \geq 1$, the the approximation factor scales as $3\alpha$. However if $\gamma(\lambda)$ is underestimated, i.e., $\alpha < 1$ then the claimed approximation guarantee does not hold in general. Our weighted $k$-center formulation has similarities to classical work on weighted $k$-center where one has a hard constraint on the total weight of the selected subset \cite{Hochbaum1986}. However existing approximation algorithms for the hard formulation are intricate and require an expensive graph squaring operation as a key step. Our soft Lagrangian based formulation allows for a much simpler and scalable algorithm that still enjoys theoretical guarantees.

\section{A parallel algorithm}
In order to scale our approach in the previous section, we next consider the setting where the algorithm has access to $m$ machines over which the computation can be distributed. For this case we propose the following parallel approximation algorithm.

\begin{algorithm}
\SetAlgoLined
 1. \textbf{Input:} Graph $G = (V, V \times V)$, hyper-parameter $\gamma (\lambda)$, $m$ machines.\\
 2. Partition the vertices onto the $m$ machines arbitrarily.\\
 3. For each machine $i$, run Algorithm~\ref{alg:weighted-k-center-improved} with $\gamma (\lambda)$ as input to get a set of $k$ centers $S_i$.\\
 4. Run Algorithm~\ref{alg:weighted-k-center-improved} again with $\gamma (\lambda)$ as input for points in $\cup_{i=1}^m S_i$ to output the final set of $k$ centers.
 \caption{A parallel approximation algorithm for the weighted $k$-center objective.}\label{alg:weighted-k-center-improved-parallel}
\end{algorithm}

\begin{theorem}
\label{thm:weighted-k-center-improved-parallel}
For a graph $G = (V, V \times V)$, let $S^*$ be the subset that optimizes the $k$-center objective in \eqref{eq:weighted-k-center-obj}. Then there exists a threshold $\gamma$ such that Algorithm~\ref{alg:weighted-k-center-improved-parallel} when run with $\gamma (\lambda)$ outputs a set $S$ of size $k$ such that 
\begin{align*}
    &\max_{i \in G} \min_{j \in S} d(i,j) +  \lambda \sum_{\hat{c}_j \in S} w(\hat{c}_j) \\
    &\leq 14 \Big(\max_{i \in G} \min_{j \in S^*} d(i,j) + \lambda \sum_{c^*_i \in S^*} w(c^*_i) \Big).
\end{align*}
\end{theorem}
\begin{proof}
Again we consider the invocation of the algorithm with $\gamma (\lambda) = \text{OPT}$ as defined in the proof of Theorem~\ref{thm:weighted-k-center-improved}.
Consider any partition $S_i$ that consists of a subset of some optimal centers in $S^*$. From the guarantee of Algorithm~\ref{alg:weighted-k-center-improved}, for each $c^*_j$ in $S_i$, either a center from another optimal cluster that is $3\gamma (\lambda)$ close to $c^*_i$ will be picked, or step $5$ will be executed with $c^*_i$ as a candidate. In the latter case, a point of weight at most $w(c^*_i)$ will be picked that is at most $2\gamma (\lambda)$ away from $c^*_i$. As a result, the set $\cup_i S_i$ will have the guarantee that there exists a subset of at most $k$ points of weight bounded by $\sum_j w(c^*_j)$ such that each $c^*_i$ is at most $3\gamma (\lambda)$ away from the subset, and as a result each point in $\cup_i S_i$ is at most $4\gamma (\lambda)$ away from the subset. By running Algorithm~\ref{alg:weighted-k-center-improved} again in $\cup_i S_i$, we will obtain a subset of $k$ centers of weight at most $\sum_j w(c^*_j)$ such that each point in $\cup_i S_i$ is at most $12\gamma (\lambda)$ away from the centers output by the algorithm. This implies that each $c^*_i$ is at most $13\gamma(\lambda)$ away from the centers. As a result we obtain a $14$-approximation to the overall objective.
\end{proof}

\noindent
{\bf Complexity.} The complexity of the running time of the sequential weighted k-center procedure (Algorithm~\ref{alg:weighted-k-center-improved}) is $O(k|V| \log(|V|))$ and matches the running time of the greedy algorithm for $k$-center~\cite{gonzalez1985clustering}. In contrast, the parallel version proposed in Algorithm~\ref{alg:weighted-k-center-improved-parallel} has a running time of $O \big(\frac{k |V| \log(|V|)}{m} + k^2 m \log(km) \big)$, where $m$ is the number of machines.

\section{Experiments}
\label{sec.exp}

\noindent
We refer to our subset selection algorithm as {\sc DUKE} to denote Diverse and Uncertain $K$-center Embeddings. In all our experiments, we use single-shot setting where we consider one budget or batch and the performance is studied for different subset selection algorithms. This is in contrast to sequential setting where multiple batches are selected and the model is retrained after every subset selection.

\begin{figure*}[!t]
    \centering
    \subfloat[]{
        \includegraphics[width=0.48\textwidth]{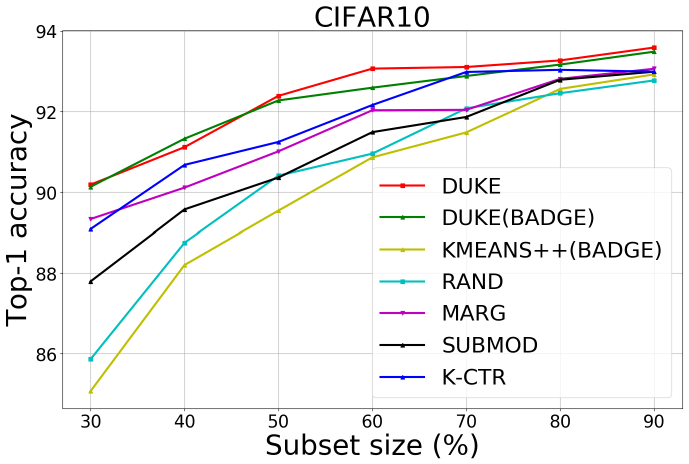}
        \label{fig:cifar10}
    }
    \subfloat[]{
        \includegraphics[width=0.48\textwidth]{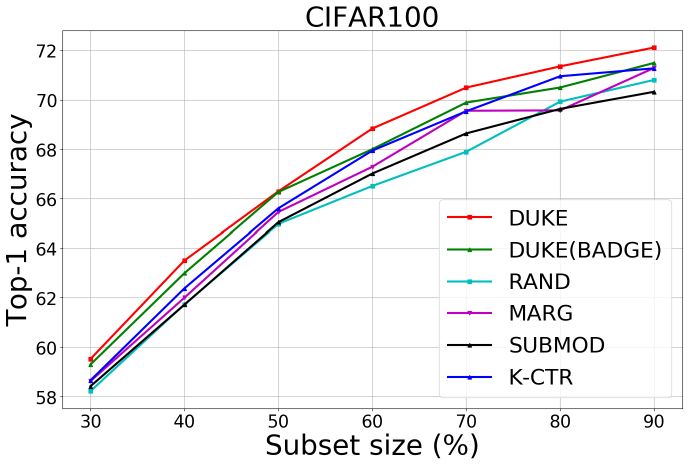}
        \label{fig:cifar100}
    }
    \\
    \subfloat[]{
        \includegraphics[width=0.48\textwidth]{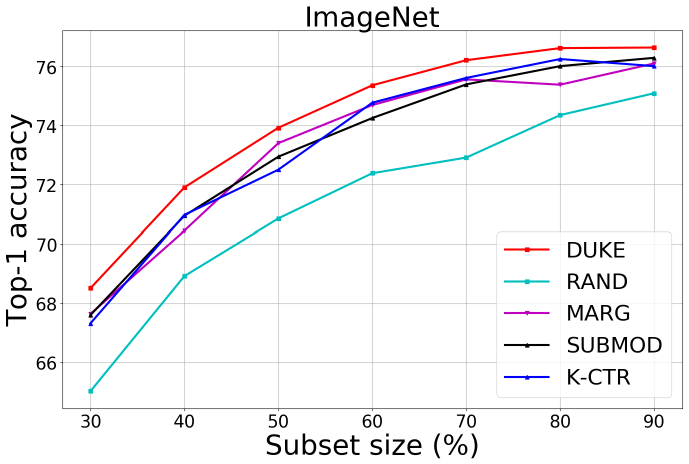}
        \label{fig:imagenet}
    }
    \subfloat[]{
        \includegraphics[width=0.48\textwidth]{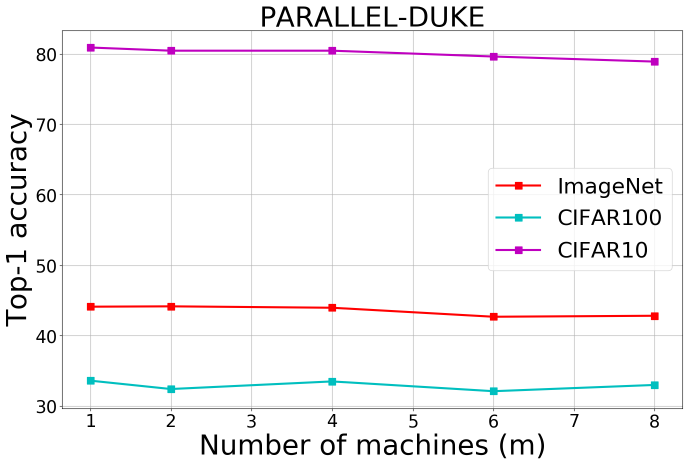}
        \label{fig:parallel}
    }
    \caption{\small \it All the Top-1 accuracy numbers are computed by averaging three trials on CIFAR-10, CIFAR-100~\cite{krizhevsky2009learning} and ImageNet~\cite{russakovsky2014imagenet}. In (d), we show the performance of our parallel algorithm with 2, 4, 6, and 8 machines.}
    \label{fig:results}
\end{figure*}

\noindent
{\bf Datasets and preprocessing.}
Standard vision benchmarks such as CIFAR-10~\cite{krizhevsky2009learning}, CIFAR-100~\cite{krizhevsky2009learning}, and ImageNet~\cite{russakovsky2014imagenet} are used for evaluation. We use a small set of points with labels to produce the initial model. In all three datasets, a random $10\%$ of the data points is used to produce the initial model. The initial model is then used to produce embeddings and class probability scores. The features of CIFAR-10, CIFAR-100, and ImageNet are of dimension $64$, $64$, and $2048$, respectively. In order to compute the distances between the data points we use the fast similarity search~\cite{avq_2020}. We construct the nearest neighbor graph $(G, E)$ by finding the $k=10$ nearest neighbors of each data point using the associated embeddings. We evaluated different distance measures such as cosine, $L_1$, and $L_2$ distances, and found no significant differences in the final performance. All the reported experiments use cosine distances. 

\noindent
{\bf Baselines.}
We show 5 subset selection methods as baselines: {\sc Rand}, {\sc Submod}~\cite{Kim2016}, {\sc Marg}~\cite{Roth2006}, {\sc $k$-ctr}~\cite{sener2017active}, and {\sc BADGE}~\cite{Ash2020Deep}.

\noindent
{\sc Rand.} We select samples from the unlabeled dataset uniformly at random.

\noindent
{\sc Marg.} Margin sampling refers to the selection of points based on the uncertainty values as given by Eq.~\ref{eq:margin_sampling}. Based on the budget $k$, we select the bottom-$k$ points that have high uncertainty or low margin scores. 

\noindent

\noindent
{\sc Submod.} We select the subset by maximizing the following submodular objective function under the budget constraint $|S| \le k$. 
\begin{equation}
\sum_{i\in S}\operatorname{utility}(i) - \lambda_{s} \sum_{i,j:~(i,j)\in E,~~i,j\in S}s(i,j),
\label{eq:submodular}
\end{equation}
This function can be shown to be monotonic submodular and an efficient, greedy algorithm can be used to identify the subset. The greedy algorithm provides a factor 2-approximation gaurantee~\cite{Nemhauser_MP1978}.
We maximize the above objective function where $\operatorname{utility}(i)$ denotes the individual utility of the different points, and $s(i,j)$ denotes the similarity between adjacent points from the nearest neighbor graph. Overall, the objective function prefers to select points with high utility, while ensuring that two points that are close in the embedding space are avoided. We tuned for best performance and selected $\lambda_s =0.9$ in our experiments. 

\noindent
{\sc k-ctr.}  ~\cite{sener2017active} propose two solutions for the approximate core-sets using $k$-center based on mixed integer programming and a greedy algorithm that comes with 2-approximation guarantees. We use the greedy algorithm for implementing ${k-ctr}$ and the computational complexity $O(k|V|\log(|V|)$ where $|V|$ is the size of the dataset and $k$ is the number of centers. Note that the computational complexity and {\sc DUKE} are the same.

\noindent
{\sc BADGE.} We use k-MEANS++ to find a set of diverse and uncertain samples using the gradients of the final layer of the network~\cite{Ash2020Deep}. The dimensions of the gradient embeddings is given by $lm$, where $l$ is the number of labels and $m$ is the dimensions of the image features (e.g., 64 for CIFAR and 2048 for ImageNet). Due to large dimensions of the gradient embeddings ($> 2M$ for ImageNet), we only show BADGE baseline for CIFAR-10. 

\noindent
{\bf Other baselines.} Subset selection is a classical problem, and there are several other baseslines one could look at. As mentioned earlier, margin sampling~\cite{Roth2006}, core-sets based on $k$-center~\cite{sener2017active}, and BADGE~\cite{Ash2020Deep} are strong baselines that are shown to outperform other methods. In particular, we also looked at other baselines such as max-entropy and variation ratios in ~\cite{Gal2017}, but the performance was similar to the margin sampling reported in this paper. In addition, we also evaluated k-median~\cite{Har-Peled2005} and it was consistently inferior to $k$-center for all the budgets. 

\noindent
{\bf Hyperparameters.}
We used ResNet-56 backbone for all the experiments. For CIFAR-10 and CIFAR-100, we used 450 epochs and the learning rate is divided by 10 at 15, 200, 300, and 400 epochs, respectively. In total, we used 90 epochs for ImageNet and the learning rate is divided by 10 at 5, 30, 60, and 80 epochs, respectively. We set the base learning rate to 1.0 for CIFAR datasets, and 0.1 for ImageNet. We used SGD with momentum 0.9 (with nestrov) for training. We report the top-1 accuracy on all these datasets. When trained with the full training set, the top-1 accuracies for CIFAR-10, CIFAR-100, ImageNet, and CIFAR-100-LT are 93.04\%,71.37\% and 76.39\%, respectively. We use the same settings for evaluating the subsets generated by the various methods. 

Some of the baselines like margin sampling, BADGE, and $k$-center do not use any hyperparameter tuning. On the other hand, the proposed algorithm {\sc DUKE} and {\sc SUBMOD} use one hyperparamter $\lambda$, and $\lambda_s$, respectively. 

{\bf Tunable hyperparameter $\lambda$.} There is only one tunable hyperparameter $\lambda$ for {\sc DUKE}. We use cosine distances between the embeddings and this typically vary from 0 to 2, and the uncertainty values vary from 0 to 1. The objective function given in Equation~\ref{eq:weighted-k-center-obj} has one distance term and the several weight terms. Since we consider different budgets varying from $30\%$ to $90\%$, we use a $\lambda$ that is dependent on the budget size to balance the influence from diversity and uncertainty. In particular, we choose $\lambda = 0.1\frac{1}{|S|}$ where $S$ is the cardinality of the selected subset. 

{\bf Choice of $\gamma$.}
Please note that the $\gamma(\lambda)$ parameter in our algorithm is not the standard hyperparameter for training and it is not tuned manually. For a given $\lambda$, we find subsets and compute the objective function in  Equation~\ref{eq:weighted-k-center-obj} for different $\gamma$'s. We choose the $\gamma$ that produces the lowest objective value for a given $\lambda$. In all our experiments, we searched over 8 $\gamma$ parameters and it is important to note that we do not train the ResNet model while selecting the right $\gamma$. This difference is crucial since the we can generate subsets for even large datasets such as ImageNet in a matter of seconds, while the actual model training takes several days. 

\noindent
{\bf Implementation.}
While there exists efficient greedy algorithms for $k$-center problem~\cite{gonzalez1985clustering, hochbaum1985best}, the proposed approach for weighted $k$-center in Algorithm~\ref{alg:weighted-k-center-improved} is not a greedy one. We show that it can still be implemented efficiently using priority queues with complexity $O(k|V| \log(|V|))$ as shown below:
\begin{algorithm}[!t]
\SetAlgoLined
 1. \textbf{Input:} Graph $G = (V, V \times V)$, hyper-parameter $\gamma$.\\
 2. Initialize an empty priority queue $pq$. \\
 3. For all $v \in V$ do $pq \leftarrow \text{ADD}(pq,(v,w(v))$ where the weight $w(v)$ is based on $v$'s uncertainty. \\
 4. Initialize $S = \{c\}$, where $(c,w) = \text{POP}(pq)$.\\
 5. For every $n \in \text{NEIGHBOR}(c)$ do $pq \leftarrow \text{REMOVE}(pq,n)$. \\
 6. While $|S| < k$ and $\text{NOT-EMPTY}(pq)$ do:
 \begin{itemize}
     \item 7. $(c,w) \leftarrow \text{POP}(pq)$.
     \item 8. $S \leftarrow S \cup c_n$, where $c_n$ has the smallest weight in $V - S$ and $|c - c_n| \le \gamma$. 
     \item 9. $pq \leftarrow \text{REMOVE}(pq_,c_n)$
     \item 10. For all $n \in \text{NEIGHBOR}(c_n)$ do $pq \leftarrow \text{REMOVE}(pq,n)$.
 \end{itemize}
 11. If $|S| < k$ do: \\
 \begin{itemize}
     \item 12. Initialize an empty priority queue pq.
     \item 13. For all $v \in V-S$ do $pq \leftarrow (v,w(v))$.
 \end{itemize}
 14. While $|S| < k$, do $S \leftarrow S \cup c$ where $(c,w) = \text{POP}(pq)$. \\
 15. Output $S$. 
 \caption{Implementation of Algorithm~\ref{alg:weighted-k-center-improved} using a priority queue.}
 \label{alg:weighted-k-center-pq-implementation}
\end{algorithm}

\noindent
{\bf Evaluation and visualization.}
We show results on different size subsets in Figure~\ref{fig:results}. {\sc DUKE} clearly outperforms the baselines such as {\sc random}, {\sc margin}, {\sc submodular}, {\sc $k$-center}, and {\sc BADGE} for smaller budgets. 
We use the same evaluation protocol used in ~\cite{ramalingam2021balancing}. 
For each subset selection method and subset size, after selecting the subset, we reveal its labels and retrain a model on the annotated subset. Accuracy of this model is then reported on the standard test set. The naive union method does better than the standard baselines, but is inferior to the proposed algorithm. Note that the naive union does not come with constant approximation gaurantees.

Both {\sc submodular} and {\sc $k$-center} compute diverse and informative subsets, and they tend to perform similar. {\sc Margin} and {\sc $k$-center} are strong baselines, and they tend to be somewhat complementary to each other. As shown in the tables, we observe that combining diversity and uncertainty does produce improvement in both the BADGE and weighted $k$-center methods. Overall, the weighted $k$-center produces more than 1\% improvement over strong baselines such as {\sc Margin}, {\sc Submodular} and {\sc $k$-center} in smaller budget subsets.

{\sc DUKE} outperforms {\sc BADGE} on CIFAR-10. We also show results that utilize gradient embeddings instead of the standard CNN features with our algorithm {\sc DUKE} in CIFAR-10 and CIFAR-100. As we can see, the proposed method {\sc DUKE}(BADGE) produces good results with gradient embeddings compared to k-MEANS++. In \cite{Ash2020Deep}, both k-MEANS++ and $k$-Determinantal Point Process were used for sampling, and we believe that {\sc DUKE} can be a beneficial alternative to k-MEANS++ and $k$-DPP. In general, we find that the gradient embeddings are useful, but k-MEANS++ leads to inferior results compared to our method. Note that the gradient embeddings have dimensions $lm$ where $l$ is the number of labels and $m$ is the dimensions of the feature embeddings. In datasets involving too many classes like ImageNet, BADGE gradient embeddings introduces huge computational bottleneck and some dimensionality reduction techniques will be necessary. 

In Figure~\ref{fig:results}(d), we validate our proposed parallel algorithm. We show the performance of 10\% subsets computed using the standard weighted $k$-center and parallel ones running on 2, 4, 6, and 8 machines respectively. While the theoretical bound we derive for the parallel algorithm uses 14-approximation, in practice we only suffer marginal loss in performance as we increase the number of machines. 

\section{Discussion}
We propose {\sc DUKE} (Diverse and Uncertain $k$-center Embeddings) using a novel and principled approach that combines two popular measures: margin sampling and $k$-center clustering. In addition to showing factor 3-approximation guarantee for the main algorithm, we prove 14-approximation guarantee for the parallel version. We consistently perform similar or superior to other baselines on vision datasets such as CIFAR-10, CIFAR-100, and ImageNet. ~\cite{Ash2020Deep} reports that diversity methods tend to do well in the initial part of the training, and uncertainty-based ones help in the later stage. We believe that some form of curriculum where we learn different $\lambda$ in Eq.~\ref{eq:weighted-k-center-obj} at different epochs during training might be a fruitful research direction.

\clearpage
%
%
\bibliographystyle{alpha}
\bibliography{main}
\end{document}